\documentclass{article} %
\usepackage{times}
\usepackage{arxiv}
\usepackage{natbib}

\usepackage[utf8]{inputenc} %
\usepackage[T1]{fontenc}    %
\usepackage{hyperref}       %
\usepackage{url}            %
\usepackage{booktabs}       %
\usepackage{amsfonts}       %
\usepackage{nicefrac}       %
\usepackage{microtype}      %
\usepackage{amsmath}
\usepackage{multirow}
\usepackage{graphicx}
\usepackage{adjustbox}
\usepackage{subfig}
\graphicspath{ {./images/} }

\title{Generated Loss and Augmented Training of MNIST VAE}

\author{
  Jason Chou\thanks{Work done during the tenure as a Google employee.} \\
  Google LLC. \\
  1600 Amphitheatre Parkway \\
  Mountain View, CA 94043 \\
  \texttt{chuanchih@gmail.com} \\
}

\begin{document}

\maketitle

\begin{abstract}
  The variational autoencoder (VAE) framework is a popular option for training unsupervised generative models, featuring ease of training and latent representation of data. The objective function of VAE does not guarantee to achieve the latter, however, and failure to do so leads to a frequent failure mode called posterior collapse. Even in successful cases, VAEs often result in low-precision reconstructions and generated samples. The introduction of the KL-divergence weight $\beta$ can help steer the model clear of posterior collapse, but its tuning is often a trial-and-error process with no guiding metrics. Here we test the idea of using the total VAE loss of generated samples (generated loss) as the proxy metric for generation quality, the related hypothesis that VAE reconstruction from the mean latent vector tends to be a more typical example of its class than the original, and the idea of exploiting this property by augmenting training data with generated variants (augmented training). The results are mixed, but repeated encoding and decoding indeed result in qualitatively and quantitatively more typical examples from both convolutional and fully-connected MNIST VAEs, suggesting that it may be an inherent property of the VAE framework.
\end{abstract}

\section{Introduction}

The variational autoencoder (VAE) framework \citep{kingma2013auto} has been a popular options for training deep generative models. It is a maximum likelihood model and establishes a lower bound (evidence lower bound, ELBO) on the likelihood of observed data by factoring out a latent representation. The VAE encoder is trained to encode training examples into posterior distributions in the latent space, and the VAE decoder is trained to reconstruct the training examples from the latent vectors sampled from these posterior distributions. To maximize ELBO, the model needs to minimize both mistakes in reconstruction (`reconstruction loss') and the difference between the posterior distributions and an assumed prior (`latent loss', measured by KL-divergence). If the latent loss dominates, posterior distributions of the latent vectors collapse to the assumed prior and the latent vector no longer carries any information about the training example (posterior collapse). Models of continuous data types in posterior collapse often hedge their bets by always outputting a weighted average of the training examples. In addition, models capable of autoregression fall back to it in posterior collapse.

A less severe but more pervasive issue of VAE is low-precision reconstructions and generated samples \citep{sajjadi2018assessing}. It has been theorized that this issue still comes down to the strength of the KL-divergence term. With strong KL-divergence term, posterior distributions of sampled latent vectors of two distinct training examples overlap and result in ambiguous and therefore low-precision generated samples. On the other hand, there may be holes in the latent space that is not covered by any posterior distributions when the KL-divergence term is weak, and such holes will result in samples unconstrained by training data \citep{rezende2018taming}. In order to steer clear of the posterior collapse and optimize posterior distributions for generation, the KL-divergence weight $\beta$ is often introduced as a hyperparameter \citep{higgins2017beta} and sometimes used in conjunction with manual annealing over training steps \citep{DBLP:journals/corr/BowmanVVDJB15,sonderby2016ladder,DBLP:journals/corr/abs-1804-02135}. While tuning of $\beta$ has shown to be effective, hyperparameter sweep over it is a laborious process and auto-tuning of it w.r.t. reconstruction error constraints has been recently proposed \citep{rezende2018taming}. It is perhaps worth pointing out a possible disconnect between the theoretical motivation and the use in practice: VAEs are designed to maximize the likelihood of training examples but often used as generative models. It is natural to wonder whether VAEs themselves find their generated sample to be a likely observation.

In this short paper we follow up \citet{anon2019generated} and 
test the following ideas with a simple convolutional MNIST VAE: 1. Measuring the generated loss, the total VAE loss of generated samples as if they were training or testing data, as a guiding metric. 2. Repeated encoding and decoding using the mean latent vector result in more typical examples of their class than the original. 3. Exploiting 2. by augmenting training data with generated variants, as a variational method for training VAEs to achieve improved generation quality. The overall results are mixed, but 2. does hold up for both convolutional and fully-connected MNIST VAE. We will draw comparisons and contrast with the original result, which focuses on discrete data.

\section{Background}

\subsection{\texorpdfstring{$\beta$}{beta}-VAE}

To recap, VAE is trained to maximize the evidence lower bound (ELBO) of the log-likelihood of training examples $x$

\begin{equation}
\log p(x) \ge \underbrace{\mathbb{E}[\log p_\theta(x|z)]}_{-\mathrm{reconstruction\ loss}} - \underbrace{KL(q_\lambda(z|x)||p(z))}_{\mathrm{latent\ loss}}
\end{equation}

where $KL(\cdot||\cdot)$ is the KL-divergence between two distributions and $z$ is the latent vector, whose prior distribution $p(z)$ is most commonly assumed to be multivariate unit Gaussian. $\log p_\theta(x|z)$ is given by the decoder, and $q_\lambda(z|x)$ is the posterior distribution of the latent vector given by the stochastic encoder, whose operation can be made differentiable through the reparameterization trick $z = \mu_\lambda(x) + \sigma_\lambda(x) \odot \epsilon$, $\epsilon \sim \mathcal{N}(0, 1)$ if $q_\lambda(z|x)$ is assumed to be a diagonal-covariance Gaussian.

A common modification to the ELBO of VAE is to add a hyperparameter $\beta$ to the KL-divergence term and use the following objective function:

\begin{equation}
\underbrace{\mathbb{E}[\log p_\theta(x|z)]}_{-\mathrm{reconstruction\ loss}} - \beta \underbrace{KL(q_\lambda(z|x)||p(z))}_{\mathrm{latent\ loss}} \label{eq:beta_vae}
\end{equation}

where $\beta$ controls the strength of the information bottleneck on the latent vector. For higher values of $\beta$, we accept lossier reconstruction, in exchange of higher effective compression ratio. In both cases, the generator $g(z)$ samples from the probability distribution given by the decoder $p_\theta(\tilde{x}|z)$ where $z$ is a random latent vector in generation time:

\begin{equation}
\tilde{x} = g(z) \sim p_\theta(\tilde{x}|z)
\end{equation}

In practice, explicit sampling is most frequently associated with discrete data like characters in a string. For continuous data types, decoder's output is often interpreted as the mean and used directly.

\subsection{Convolutional MNIST VAE setup}
\label{sec:cvae_setup}
In intentional contrast with \citet{anon2019generated}, we apply its ideas to a simple case with off-the-shelf models. We use the original 60000:10000 training:testing split of the MNIST dataset \citep{lecun-mnisthandwrittendigit-2010} with only the minimum preprocessing of converting the integers to floats and scaling them down by a factor 255 to make sure that they fall in the range of [0, 1]. The architecture of the convolutional VAE is taken from \href{https://www.tensorflow.org/alpha/tutorials/generative/cvae}{the TensorFlow 2.0 convolutional VAE example (accessed 2019-04-16)}, with 2 covolutional layers + 1 fully connected layer for the encoder and 1 fully connected layer + 3 convolution layers for the decoder. Cross-entropy loss is used for the reconstruction loss since both the input and reconstruction give pixel values in the range of [0, 1], and the latent loss is the usual KL-divergence. The model is trained with the Adam optimizer \citep{DBLP:journals/corr/KingmaB14} with constant learning rate $10^{-3}$. To make sure that epochs line up, we use batch size = 50 and run a testing step every 6 training steps. We also run a `generation' step every 6 training steps that generates 50 samples and measure their total VAE loss as if they were testing data. For generation, we just sample from the 32-dim unit Gaussian prior $\mathcal{N}(0, 1)$ and feed the random latent vectors to the decoder. The training budget is fixed at 5 epochs = 6000 steps, and we report the average losses over the last epoch for each experiment. The value of $\beta$ is again reported as the relative weight of the average KL-divergence loss per latent dimension to the average cross-entropy loss per pixel value.

\subsection{Fr\'echet Inception Distance and p-value}
\label{FID_and_p_value}
In order to quantify generation quality of the MNIST VAE, we use the widely adopted Fr\'echet Inception Distance (FID) \citep{DBLP:journals/corr/HeuselRUNKH17} computed with  \href{https://github.com/tensorflow/tensorflow/tree/master/tensorflow/contrib/gan}{the TensorFlow-GAN library's default graph (accessed 2019-04-17)}. With the simplifying assumption that logit activations of a trained classifier follow multivariate Gaussian distribution $\mathcal{N}(m,\Sigma)$ when given training examples and $\mathcal{N}(m_w,\Sigma_w)$ when given generated samples, FID is the Fr\'echet distance (also known as Wasserstein-2 distance) between these two distributions:

\begin{equation}
d^2=\Vert m-m_w\Vert_2^2 +\mathrm{Tr}(\Sigma+\Sigma_w-2(\Sigma^{1/2}\Sigma_w\Sigma^{1/2})^{1/2})
\end{equation}

With the same assumption, we can also quantify how `typical' a generated sample is by p-value, which can be obtained by applying the multivariate version of the two-tailed t test to its logit activations $x$ \citep{anon2019generated}:

\begin{align*}
d_m^2 &= (x - m)^T \Sigma^{-1} (x - m) \\
p &= 1 - \operatorname{CDF}_{\chi_k^2}(d_m^2)
\end{align*}

where $d_m^2$ is the Mahalanobis distance squared, $\operatorname{CDF}_{\chi_k^2}$ is the cumulative distribution function (CDF) of chi-squared distribution with $k=10$ degrees of freedom. Intuitively, p-value is the probability that the given $x$ is more likely than a sample drawn from the distribution and measures how close $x$ is to the mode of the distribution. As we will see later, however, our simplifying assumption breaks down when we try to quantify how typical a generated sample is by p-value due to a fundamental discrepancy: logit activations of a MNIST classifier follow a 10-modal distribution instead of an unimodal multivariate Gaussian distribution. Therefore, the distance from $x$ to the mean/mode of the distribution $m$ bears no relation to how typical the generated sample is as a handwritten digit. To remedy this, we make the weaker assumption that logit activations of a trained classifier follow multivariate Gaussian distribution $\mathcal{N}(m_L,\Sigma_L)$ when given training examples of a given class $L \in \{0, 1, 2, \dots, 9\}$, and calculate the conditional p-value of a sample's logit activations $x$ instead:

\begin{align*}
d_m^2 &= (x - m_L)^T \Sigma_L^{-1} (x - m_L) \\
p &= 1 - \operatorname{CDF}_{\chi_k^2}(d_m^2)
\end{align*}

where $L = \operatorname*{argmax}_i x_i$, i.e. the label given by the classifier. To gather meaningful statistics, we always evaluate FID and aggregate properties of p-values over 10000 samples.

\section{Convolutional \texorpdfstring{$\beta$}{beta}-VAE result}

\begin{figure}[ht]
\centering
\caption{Training, testing, and generated loss over the value of $\beta$, averaged over the last epoch. Values of FID were calculated over 10000 samples generated by the final model.}\label{fig:beta_loss_convol}
\includegraphics[width=\textwidth]{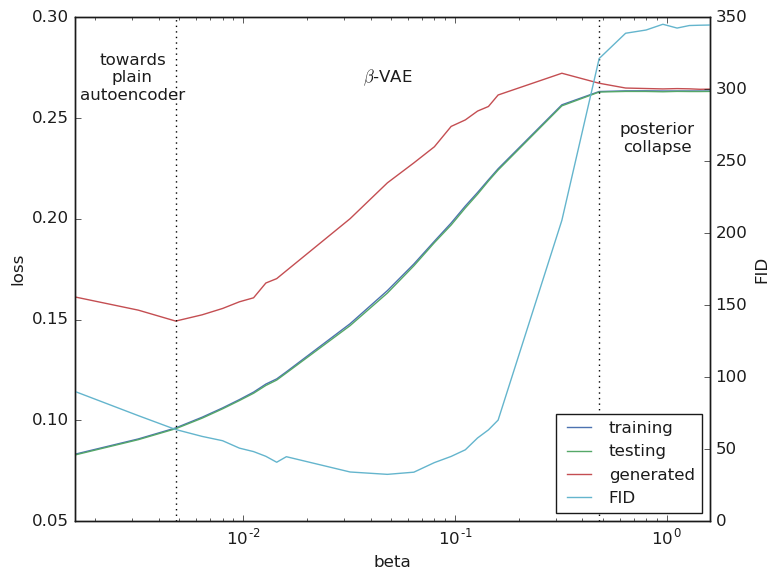}
\end{figure}

\begin{figure}[ht]
\centering
\caption{The value of $\beta$ and 10 generated samples for each model.}\label{fig:mnist_grid_convol}
\includegraphics[width=0.6\textwidth]{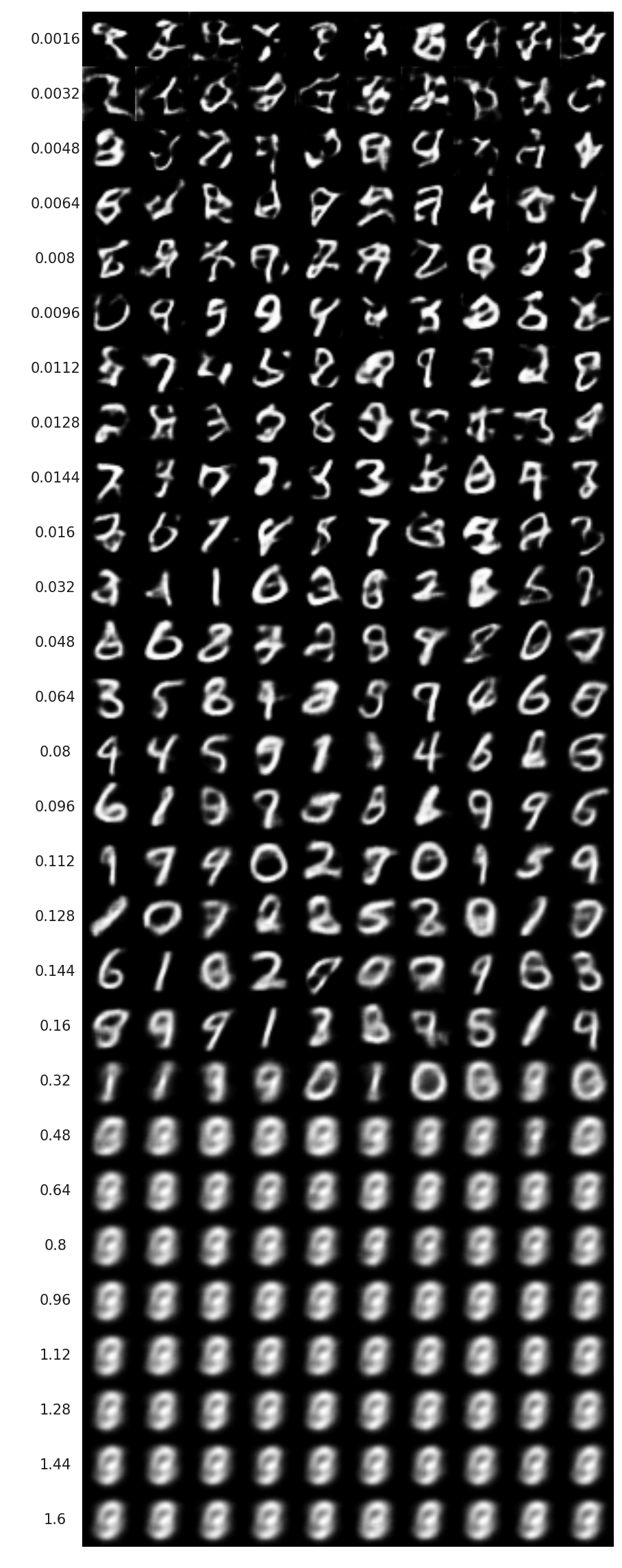}
\end{figure}

With the setup described in Sec~\ref{sec:cvae_setup}, here we report the losses of the baseline $\beta$-VAE models over the full hyperparameter sweep of $\beta$ (Fig~\ref{fig:beta_loss_convol}), and 10 generated samples for each model annotated with its $\beta$ value (Fig~\ref{fig:mnist_grid_convol}).

We again observe no overfitting, with training loss practically identical to the testing loss. Generated loss lags behind in the $\beta$-VAE regime of the hyperparameter space, but not as severely as in \citet{anon2019generated}. In the case of image VAE, both the encoder and decoder are typically continuous functions that map the pixel values to the latent vector and back. Perhaps the continuity and the magnitudes of the derivatives of the functions limit how different generated loss can be from the training/testing loss. As expected, training/testing loss decreases monotonically as we lower the weight of the latent loss, but the value of generated loss starts rising again as the $\beta$-VAE transitions towards a plain autoencoder. In the hyperparameter regime of $\beta < 0.0048$, generated samples are increasingly unconstrained by the training examples, and the model is simply not trained to autoencode them. On the other end, as the value of $\beta$ increases past the point of posterior collapse, the model starts to hedge its bet and always outputs a weighted average of the training examples, so the values of training/testing loss and generated loss converge. In this sense, the value of generated loss does delimit different regimes of the hyperparameter space. However, it is not as sensitive to the model's generation quality as FID: the value of $\beta = 0.048$ where FID is at the minimum is a plausible optimal value in terms of subjective assessment of Fig~\ref{fig:mnist_grid_convol}, but the value of $\beta = 0.0048$ where generated loss is at the minimum is clearly too low. We notice that the value of generated loss is far more sensitive to the value of $\beta$ for a shallower MNIST VAE built with only fully-connected layers (Appendix \ref{appendix:fc_vae}), so the sensitivity of generated loss may depend on the absence of built-in assumptions about the data (translational invariance in this case) in the model.

\begin{figure}[ht]
\centering
\caption{10 samples generated by the $\beta = 0.048$ model over repeated encoding and decoding using the mean latent vector, annotated with the number of repetitions.}\label{fig:0048_convol_repeated_autoencoding}
\includegraphics[width=0.5\textwidth]{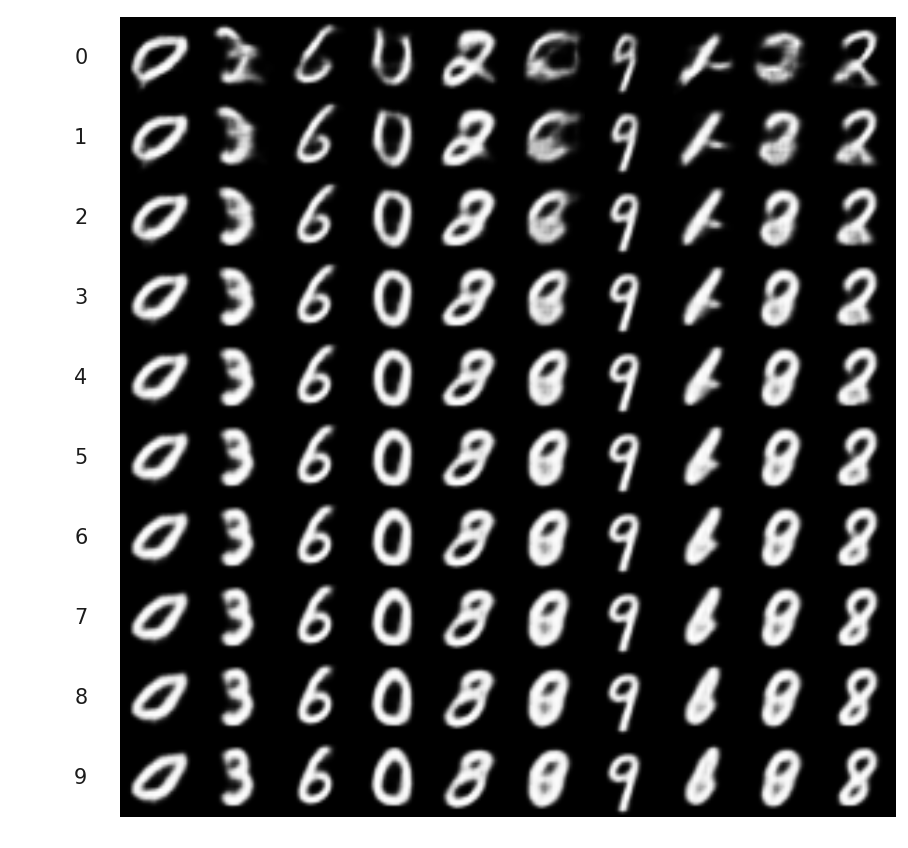}
\end{figure}

\begin{figure}[ht]
\centering
\caption{10 training examples over repeated encoding and decoding by the $\beta = 0.048$ model using the mean latent vector, annotated with the number of repetitions.}\label{fig:0048_convol_train_repeated_autoencoding}
\includegraphics[width=0.5\textwidth]{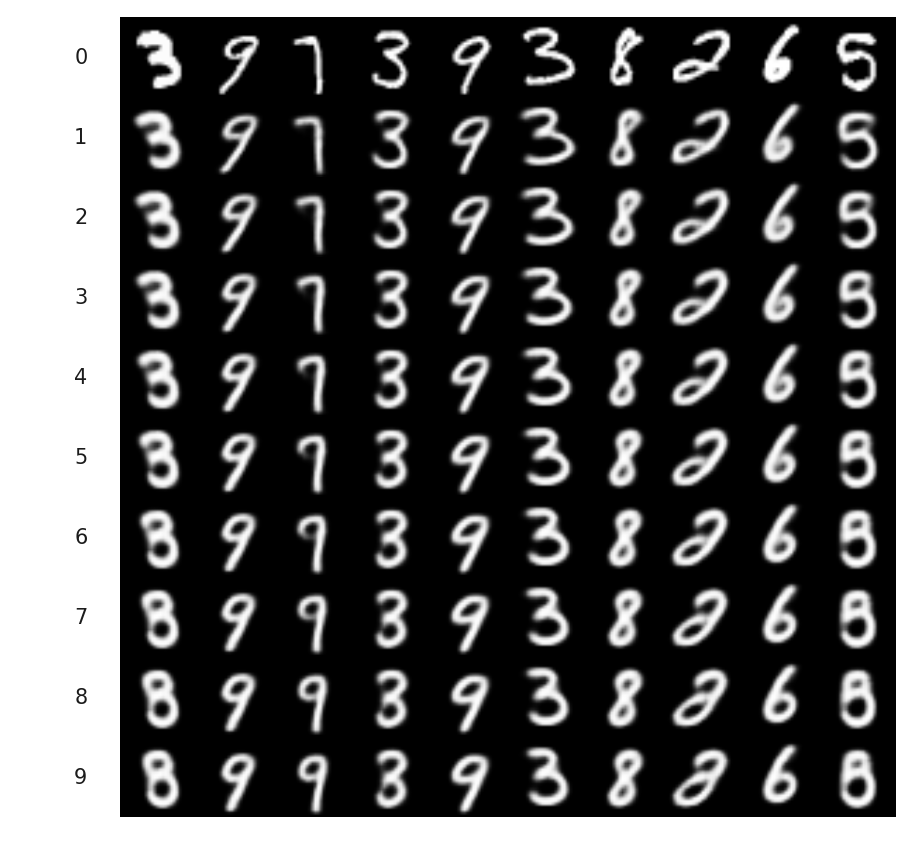}
\end{figure}

\begin{figure}[ht]
\centering
\caption{Box plot of conditional p-values over repeated encoding and decoding. The `generated' sequence at repetition = $n$ corresponds to the p-value distribution of 10000 generated samples, and the `reconstructed' sequence at repetition = $n$ corresponds to that of 10000 randomly selected training examples using the $\beta = 0.048$ model.}\label{fig:0048_convol_generated_train_repeated_conditional_p_values}
\includegraphics[width=\textwidth]{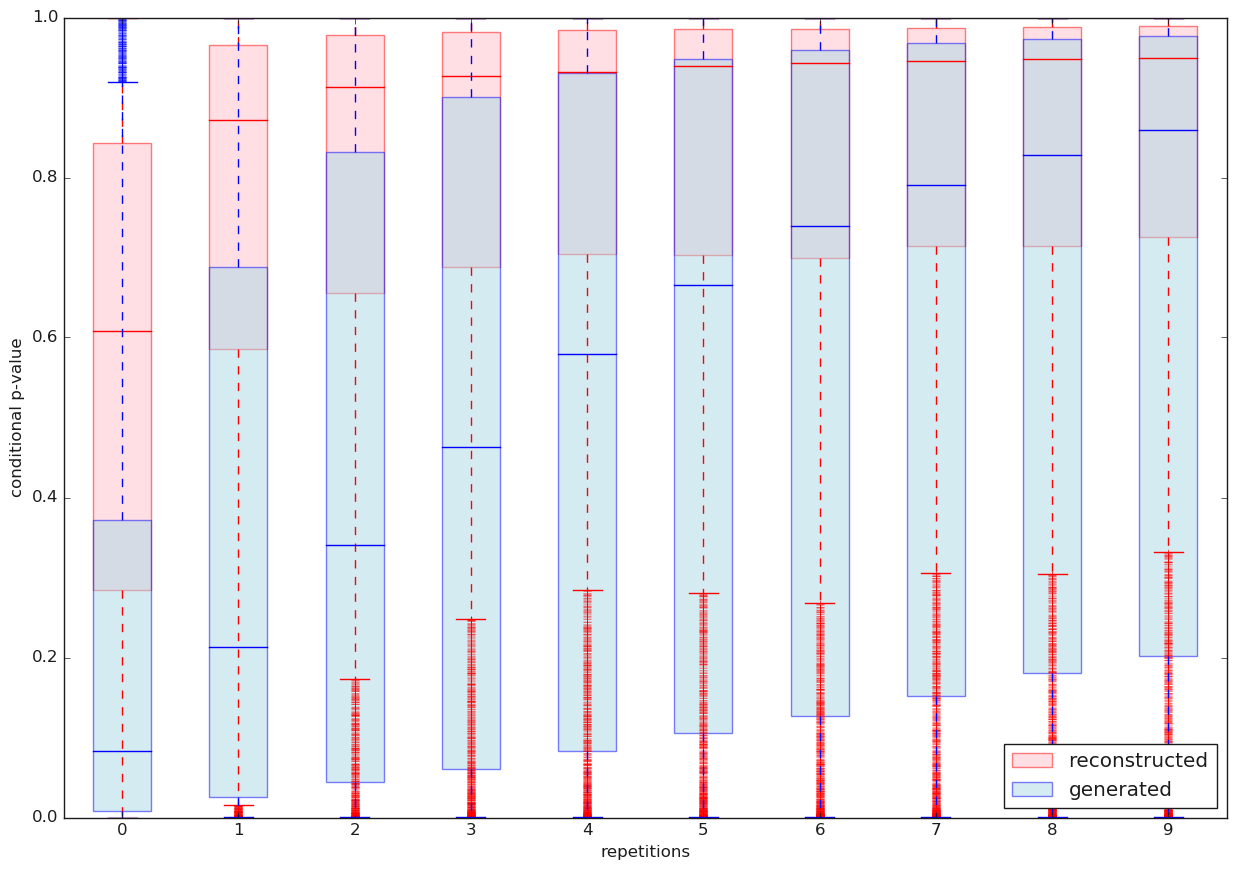}
\end{figure}

With generated loss measured, we then test the related hypothesis that VAE reconstruction from the mean latent vector tends to be a more typical example of its class. Intuitively, if this hypothesis is true, we expect reconstruction loss of generated samples (which are more likely to be atypical) to stay elevated and contribute to elevated generated loss. We again test this hypothesis by repeated encoding and decoding using the mean latent vector. Subjectively, both generated samples and training examples do seem to converge to `typical', textbook examples of handwritten digits (Fig~\ref{fig:0048_convol_repeated_autoencoding} and~\ref{fig:0048_convol_train_repeated_autoencoding}). To quantify how `typical' generated samples and training examples are over repeated encoding and decoding, we turn to p-values as described in Sec~\ref{FID_and_p_value}. The unimodal, unconditional p-value results are puzzling (Appendix \ref{appendix:unconditional_p_value}), until one realizes that classifier logit activations actually follow a 10-modal distribution and computes the conditional p-values instead (Fig~\ref{fig:0048_convol_generated_train_repeated_conditional_p_values}). The conditional p-values over repeated encoding and decoding follow the familiar pattern that both generated samples and training examples converge to the mode of the distribution and that training examples converge faster than the generated samples. Disappointingly, augmented training does not seem to be able to exploit this and improve the generation quality of the described convolutional VAE (Appendix \ref{appendix:augmented_training}).

\clearpage

\section{Conclusion and discussion}

We applied the idea of generated loss and augmented training to MNIST VAEs and obtained mixed results. More specifically,

\begin{enumerate}
  \item The total VAE loss of generated samples (generated loss) lags behind the training/testing loss but does not seem to reflect the model's generation quality, especially for convolutional models.
  \item For both convolutional and fully-connected MNIST VAEs (Appendix \ref{appendix:fc_vae}), repeated encoding and decoding using the mean latent vector again lead to more typical examples of their class.
  \item Disappointingly, augmented training is not able to exploit 2. and improve the generation quality of the given MNIST VAEs.
\end{enumerate}

In contrast to \citet{anon2019generated}, generated loss does not seem to reflect the model's generation quality and augmented training does not seem to help. For continuous VAEs such as MNIST VAE, the encoder $\mu_\lambda(x)$ and decoder $p_\theta(x|z)$ are both continuous functions carefully designed such that the derivatives are bounded. In particular, models like convolutional VAE already have built-in assumptions about the training data (translational invariance in this case). Both factors seem to allow convolutional VAE to generalize better and limit the difference between training/testing loss and generated loss. With less difference to exploit, augmented training is also no longer able to improve the model's generation quality and perhaps more sophisticated approach is necessary. The observation that repeated encoding and decoding using the mean latent vector lead to more typical examples of the class still holds, however, both qualitatively and quantitatively. It may well be an inherent property of the VAE framework useful for clustering purpose a la mean-shift clustering, given that it is tied to the variational method of minimizing reconstruction loss to the given training example from the surrounding latent space.

\bibliography{ms}
\bibliographystyle{ms}

\appendix
\section{Unconditional p-values over repeated encoding and decoding}
\label{appendix:unconditional_p_value}

\begin{figure}[ht]
\centering
\caption{Box plot of (unconditional) p-values over repeated encoding and decoding. The `generated' sequence at repetition = $n$ corresponds to the p-value distribution of 10000 generated samples, and the `reconstructed' sequence at repetition = $n$ corresponds to that of 10000 randomly selected training examples using the $\beta = 0.048$ model. In retrospect, maybe we should expect this result: Compared to the modes of the conditional distributions, logit activations of the training examples follow the true distribution and tend to be further away from the unconditional global mean $m$, whereas logit activations of the samples generated by drawing from the assumed prior $\mathcal{N}(0, 1)$ tend to be closer to the unconditional global mean $m$. Over repeated encoding and decoding, both converge to the modes of the conditional distributions but from opposite directions. }\label{fig:0048_convol_generated_train_repeated_p_values}
\includegraphics[width=\textwidth]{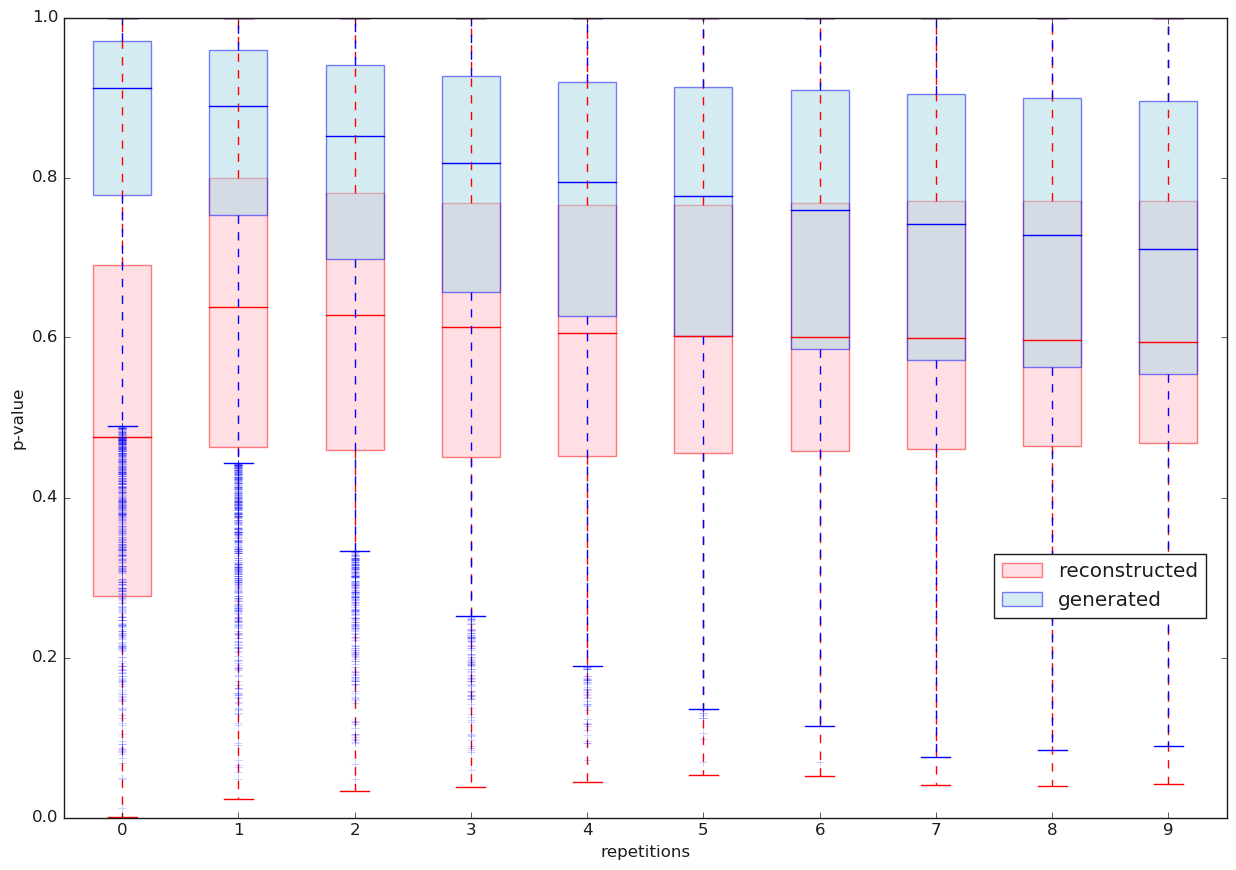}
\end{figure}

\clearpage

\section{Augmented training}
\label{appendix:augmented_training}

Augmented training is motivated by the persistent gap between generated loss and training/testing loss \citep{anon2019generated}. By augmenting the training data with generated variants, it makes sure that information from the training examples propagates beyond their respective posterior distributions. To recap, it features the  following training scheme:

\begin{adjustbox}{center}
\framebox{
\parbox{\textwidth}{
After \texttt{gen\_start\_step} steps:
\begin{enumerate}
  \item Initialize $n_{\mathrm{augmented}}$ augmented latent vectors with sampled latent vectors of the current training batch.
  \item Augment next training batch with variants generated from the augmented latent vectors.
  \item After a training step, each augmented latent vector is replaced with either:
  \begin{enumerate}
    \item The sampled latent vector of an example from the current training batch, selected without replacement, with probability $p_{\mathrm{sampled}}$.
    \item The sampled latent vector of the variant generated from it with probability $(1-p_{\mathrm{sampled}})$.
  \end{enumerate}
  \item Repeat from 2.
\end{enumerate}
}
}
\end{adjustbox}

Augmented training extends the standard VAE training scheme. Right after training to minimize reconstruction loss from the sampled latent vector, we actually generate a reconstruction from it and augments the next training batch with the reconstruction. The next training step will encode this reconstruction into its own posterior distribution in the latent space and minimize the reconstruction loss from the augmented latent vector to this reconstruction, and so on. The similarity of the successive reconstructions will decay over repeated encoding / decoding due to the model's capacity limit and the noise introduced by latent vector sampling, so we re-initialize it with probability $p_{\mathrm{sampled}}$ such that the average lifetime is $\frac{1}{p_{\mathrm{sampled}}}$ steps. We only start augmented training after \texttt{gen\_start\_step} steps to make sure that the model is already trained to generate reasonable reconstructions, and $n_{\mathrm{augmented}}$ controls the number of augmented latent vectors we use. Formally, we train the model with reconstructions from the following sequence in addition to the training examples $x$:

\begin{align*}
x' = g(z), z &\sim \mathcal{N}(\mu_\lambda(x), \sigma_\lambda(x)) \\
x'' = g(z'), z' &\sim \mathcal{N}(\mu_\lambda(x'), \sigma_\lambda(x'))  \\
&\dots \\
x^{(n)} = g(z^{(n-1)}), z^{(n-1)} &\sim \mathcal{N}(\mu_\lambda(x^{(n-1)}), \sigma_\lambda(x^{(n-1)}))  \text{ for } n > 0 \\
\end{align*}

In terms of objective function, we have

\begin{equation}
\label{eq:augmented}
\sum_{n=0}^{\infty} p^{\min(n, 1)}_{\mathrm{sampled}} (1 - p_{\mathrm{sampled}})^{\max(n-1, 0)} (\mathbb{E}[\log p_\theta(x^{(n)}|z^{(n)})] - \beta KL(q_\lambda(z^{(n)}|x^{(n)})||p(z^{(n)})))
\end{equation}

\begin{figure}[ht]
\centering
\caption{Training, testing, and generated loss over the value of $\beta$, averaged over the last epoch. Values of FID were calculated over 10000 samples generated by the final model (\texttt{gen\_start\_step} = 2400, $p_{\mathrm{sampled}} = 1$).}\label{fig:beta_loss_convol_augmented_late_1}
\includegraphics[width=\textwidth]{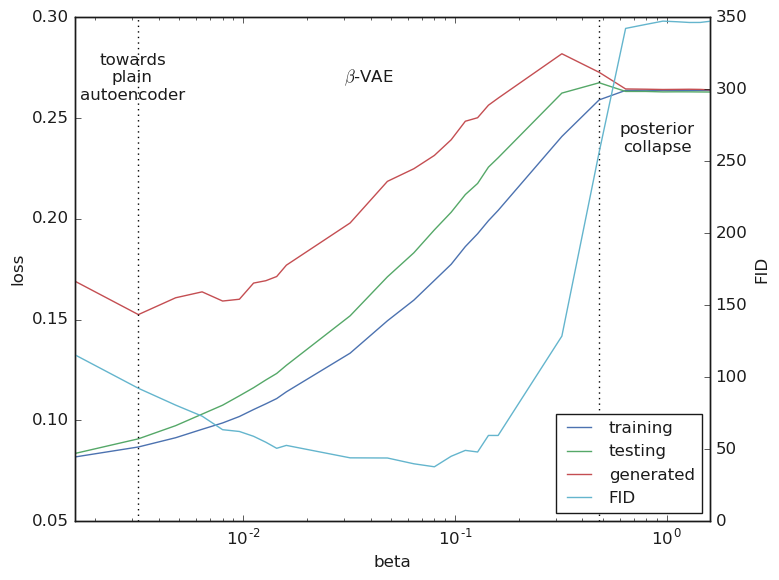}
\end{figure}

\begin{figure}[ht]
\centering
\caption{The value of $\beta$ and 10 generated samples for each model (\texttt{gen\_start\_step} = 2400, $p_{\mathrm{sampled}} = 1$).}\label{fig:mnist_grid_convol_augmented_late_1}
\includegraphics[width=0.6\textwidth]{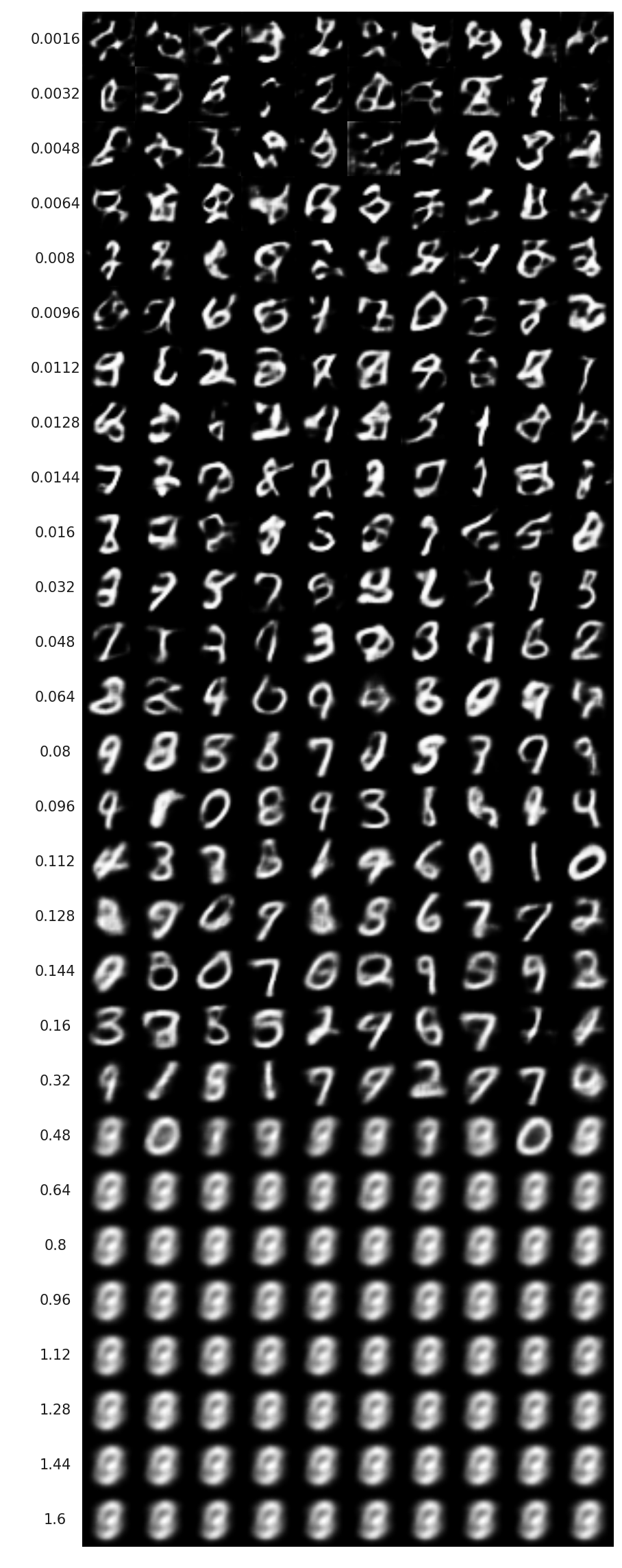}
\end{figure}

\begin{figure}[ht]
\centering
\caption{Mean conditional p-values calculated over 10000 samples generated by the baseline convolutional models vs. the augmented training convolutional models (\texttt{gen\_start\_step} = 2400, $p_{\mathrm{sampled}} = 1$).}\label{fig:convol_mean_p_values}
\includegraphics[width=0.5\textwidth]{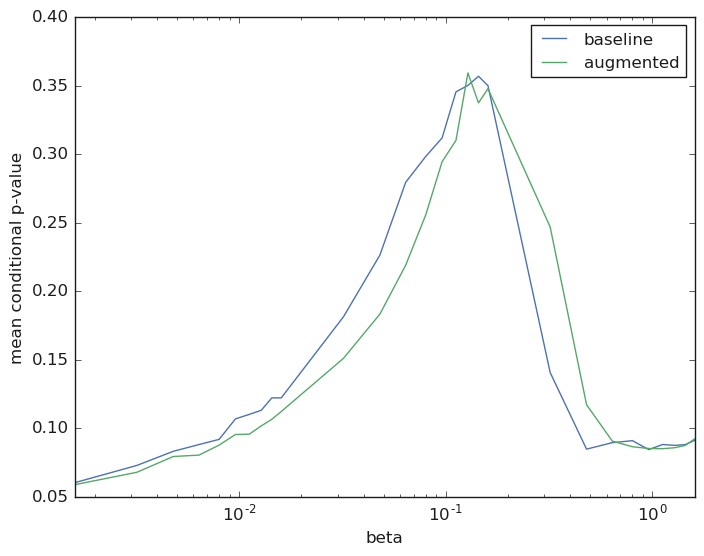}
\end{figure}

\begin{figure}[ht]
\centering
\caption{Median conditional p-values calculated over 10000 samples generated by the baseline convolutional models vs. the augmented training convolutional models (\texttt{gen\_start\_step} = 2400, $p_{\mathrm{sampled}} = 1$).}\label{fig:convol_median_p_values}
\includegraphics[width=0.5\textwidth]{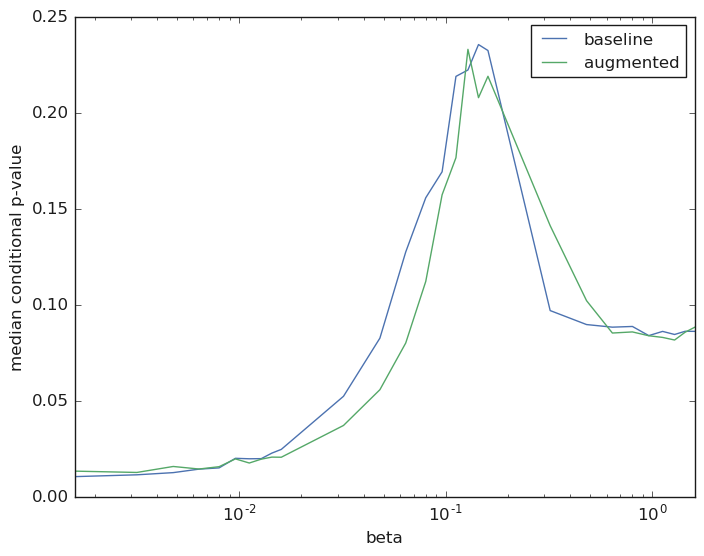}
\end{figure}

Assuming that $n_{\mathrm{augmented}}$ is equal to the training batch size, as is the case for our experiments. Disappointingly, augmented training does not seem to improve the generation quality of the described convolutional VAE, either in terms of FID or subjective evaluation (Fig~\ref{fig:beta_loss_convol_augmented_late_1} and~\ref{fig:mnist_grid_convol_augmented_late_1}, both taken from the best experiments so far). Conditional p-value statistics indicate that augmented training does not help the model generate more `typical' examples of the class, either (Fig~\ref{fig:convol_mean_p_values} and~\ref{fig:convol_median_p_values}). One may notice that generated loss is not significantly lowered by augmented training, which may serve as an indication of the case that augmented training does not help.

\clearpage

\section{Fully-connected MNIST VAE}
\label{appendix:fc_vae}

The fully-connected MNIST VAE architecture is taken from \href{https://www.tensorflow.org/alpha/guide/keras/custom_layers_and_models#putting_it_all_together_an_end-to-end_example}{the TensorFlow 2.0 Keras example (accessed 2019-04-16)}. Both the encoder and decoder consist of two fully-connected layers, with 64 intermediate dimensions and 32 latent dimensions. The first layer of the encoder is a $784 \times 64$ ReLU layer, which leads to a split second layer that uses a linear $64 \times 32$ network to generate the mean vector and another $64 \times 32$ linear network to generate the log-variance vector. The usual reparameterization trick is then used to generate the sampled latent vector from the diagonal-covariance Gaussian. The decoder takes the sampled latent vector and applies a $32 \times 64$ ReLU layer, followed by a $64 \times 784$ sigmoid layer to generate $28^2$ floats in the range of [0, 1]. The rest of the setup remains the same as described in Sec~\ref{sec:cvae_setup}.

\begin{figure}[ht]
\centering
\caption{Training, testing, and generated loss over the value of $\beta$, averaged over the last epoch. Regimes of the hyperparameter space that approach plain autoencoder, $\beta$-VAE, and posterior collapse are distinct and clearly visible.}\label{fig:beta_loss}
\includegraphics[width=\textwidth]{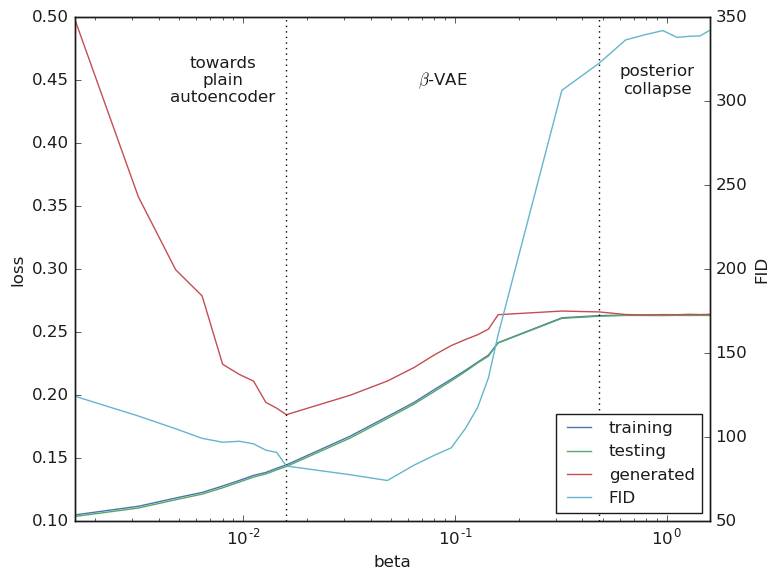}
\end{figure}

\begin{figure}[ht]
\centering
\caption{The value of $\beta$ and 10 generated samples for each model.}\label{fig:mnist_grid}
\includegraphics[width=0.6\textwidth]{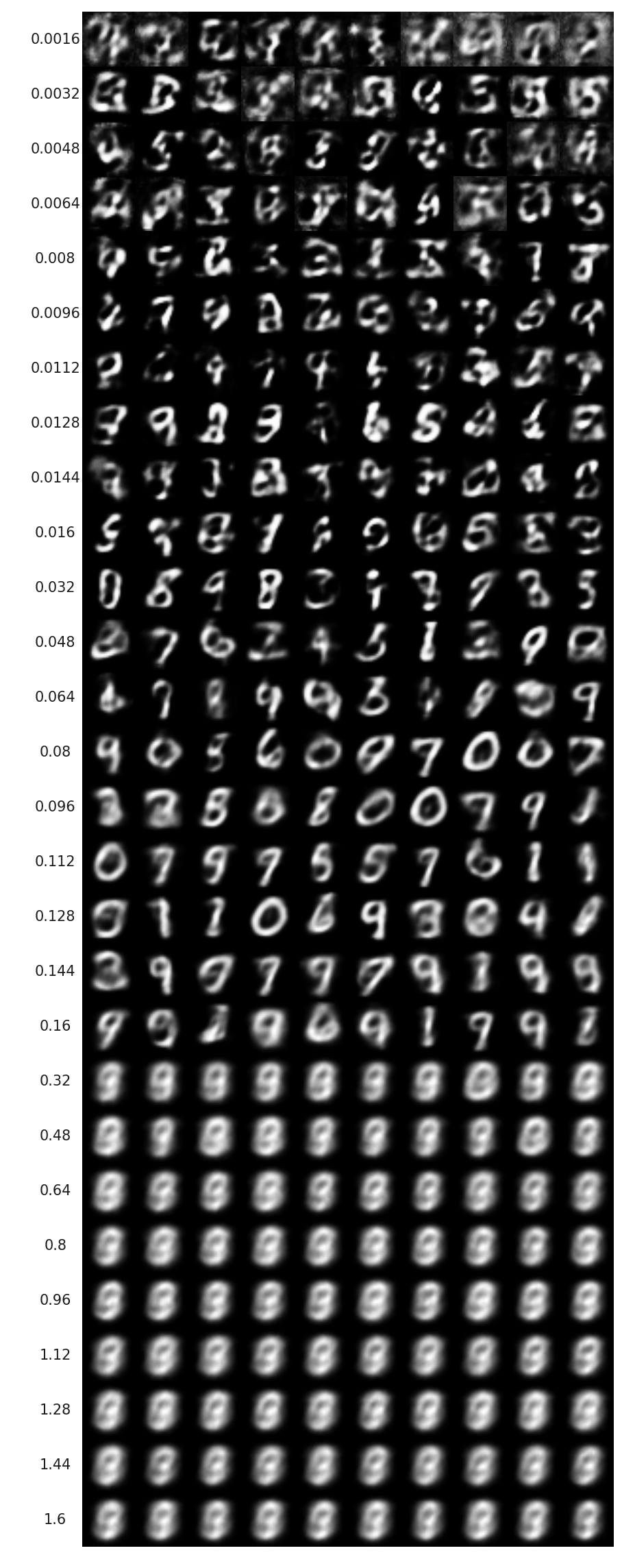}
\end{figure}

With the setup described above, here we report the losses of the fully-connected $\beta$-VAE models over the full hyperparameter sweep of $\beta$ (Fig~\ref{fig:beta_loss}), and 10 generated samples for each model annotated with its $\beta$ value (Fig~\ref{fig:mnist_grid}).

\begin{figure}[ht]
\centering
\caption{10 samples generated by the $\beta = 0.048$ model over repeated encoding and decoding using the mean latent vector, annotated with the number of repetitions.}\label{fig:0048_repeated_autoencoding}
\includegraphics[width=0.5\textwidth]{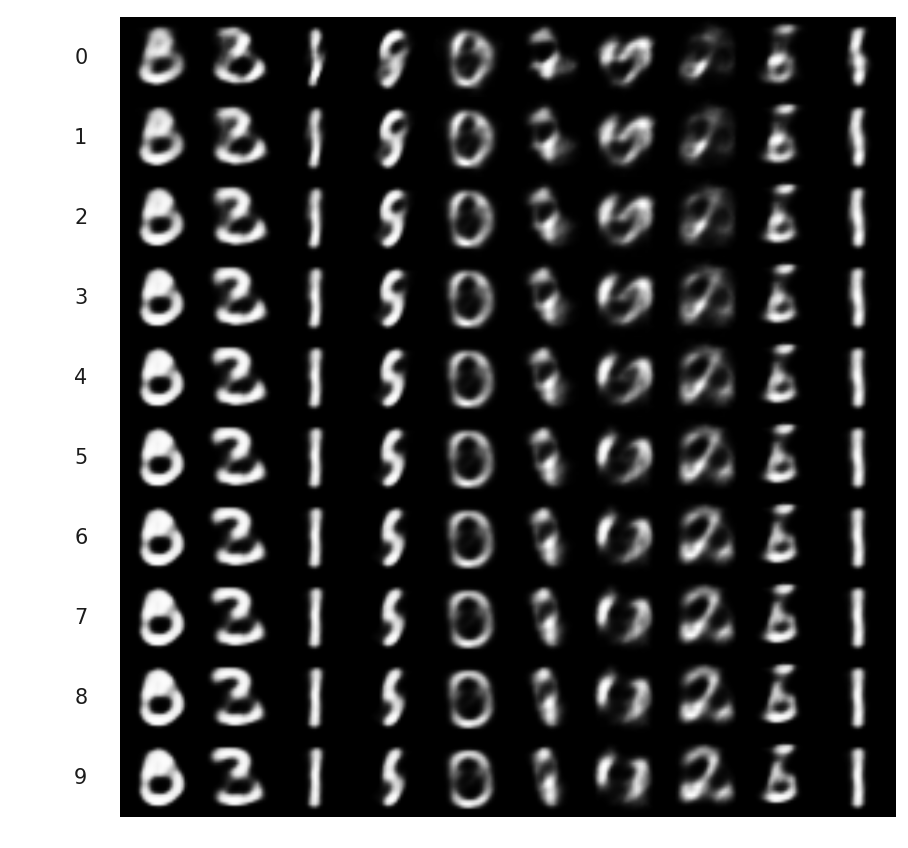}
\end{figure}

\begin{figure}[ht]
\centering
\caption{10 training examples over repeated encoding and decoding by the $\beta = 0.048$ model using the mean latent vector, annotated with the number of repetitions.}\label{fig:0048_train_repeated_autoencoding}
\includegraphics[width=0.5\textwidth]{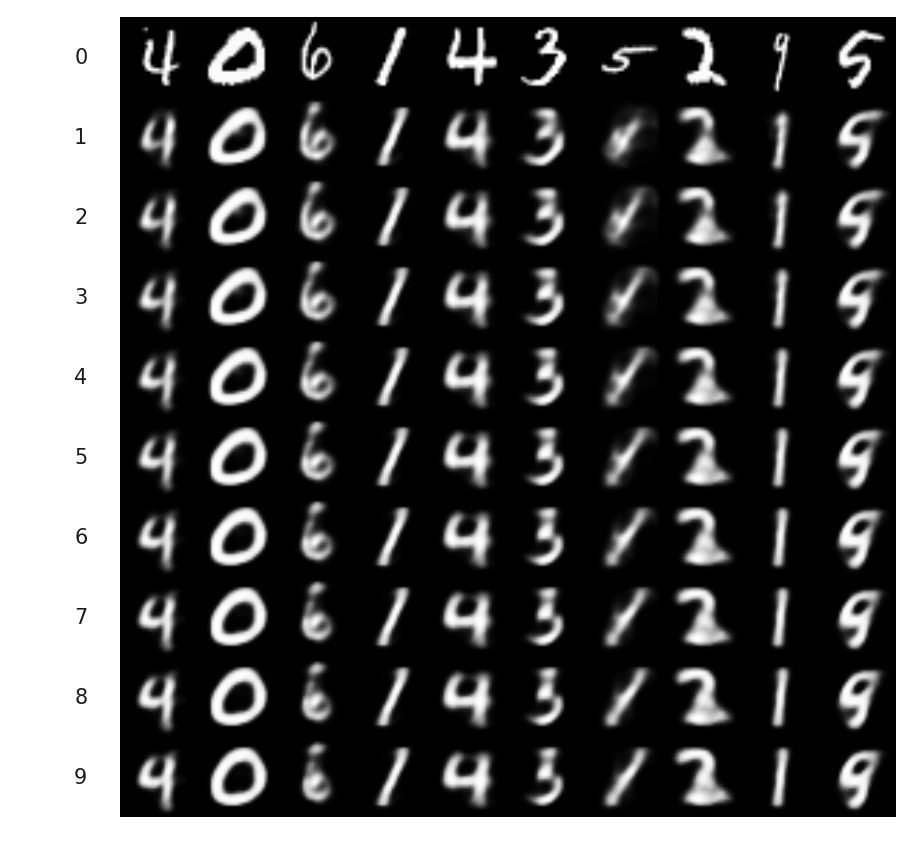}
\end{figure}

\begin{figure}[ht]
\centering
\caption{Box plot of conditional p-values over repeated encoding and decoding using the $\beta = 0.048$ model.}\label{fig:0048_generated_train_repeated_conditional_p_values}
\includegraphics[width=\textwidth]{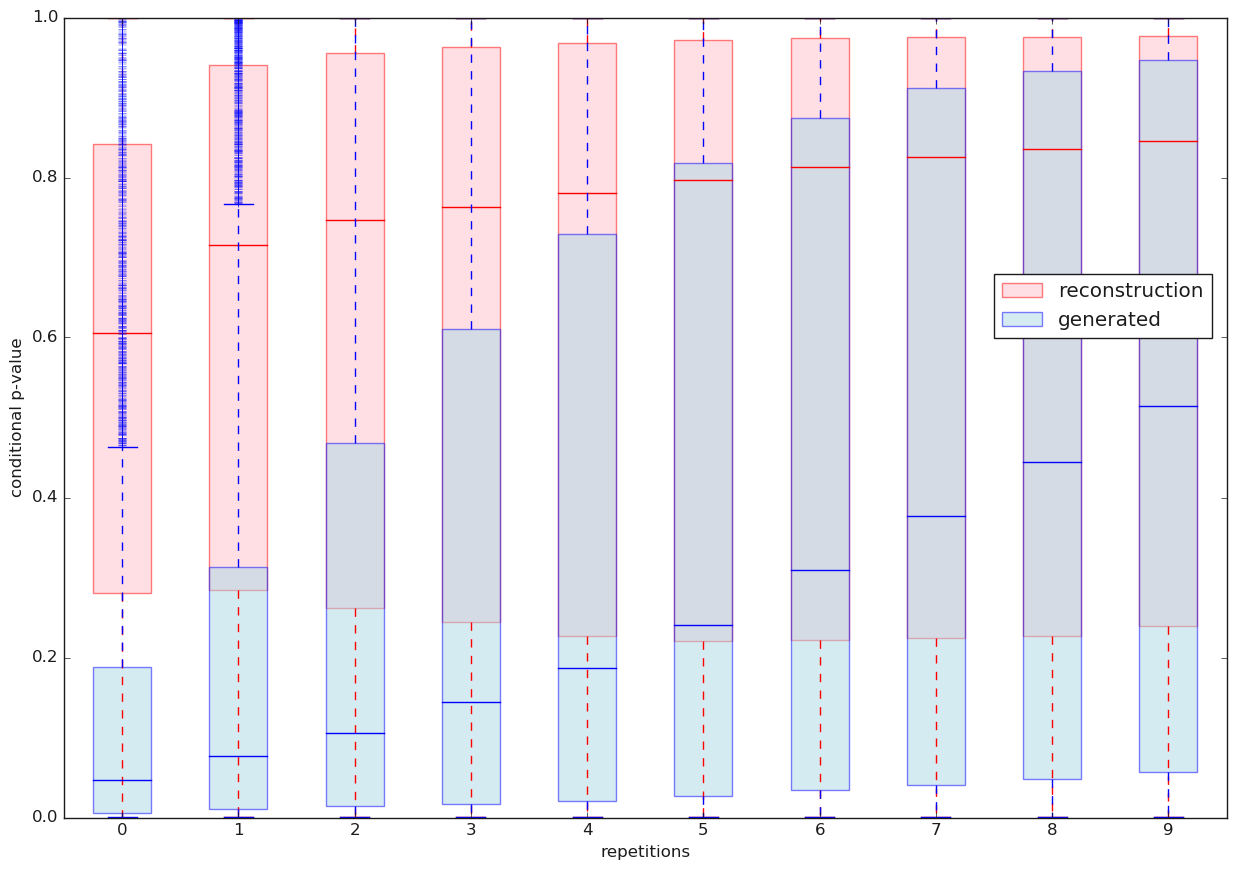}
\end{figure}

\begin{figure}[ht]
\centering
\caption{Training, testing, and generated loss over the value of $\beta$, averaged over the last epoch. Values of FID were calculated over 10000 samples generated by the final model (\texttt{gen\_start\_step} = 2400, $p_{\mathrm{sampled}} = 0.125$).}\label{fig:beta_loss_augmented_late_8}
\includegraphics[width=\textwidth]{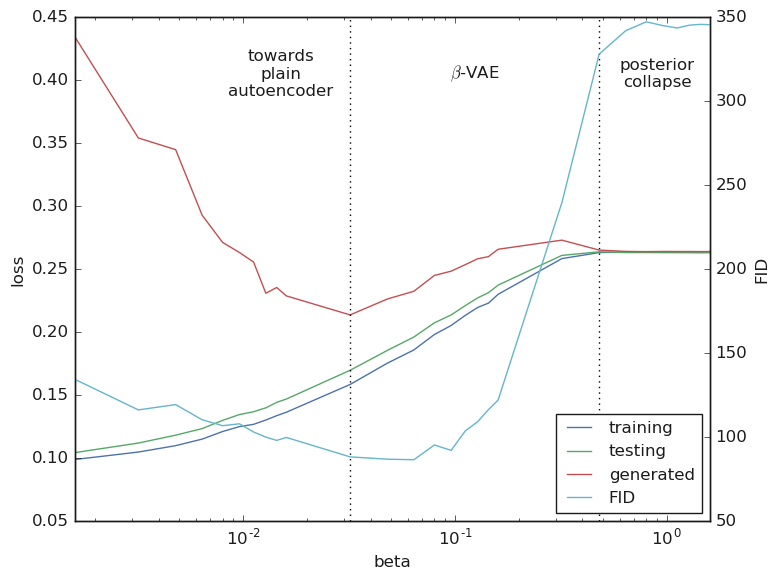}
\end{figure}

\begin{figure}[ht]
\centering
\caption{The value of $\beta$ and 10 generated samples for each model (\texttt{gen\_start\_step} = 2400, $p_{\mathrm{sampled}} = 0.125$).}\label{fig:mnist_grid_augmented_late_8}
\includegraphics[width=0.6\textwidth]{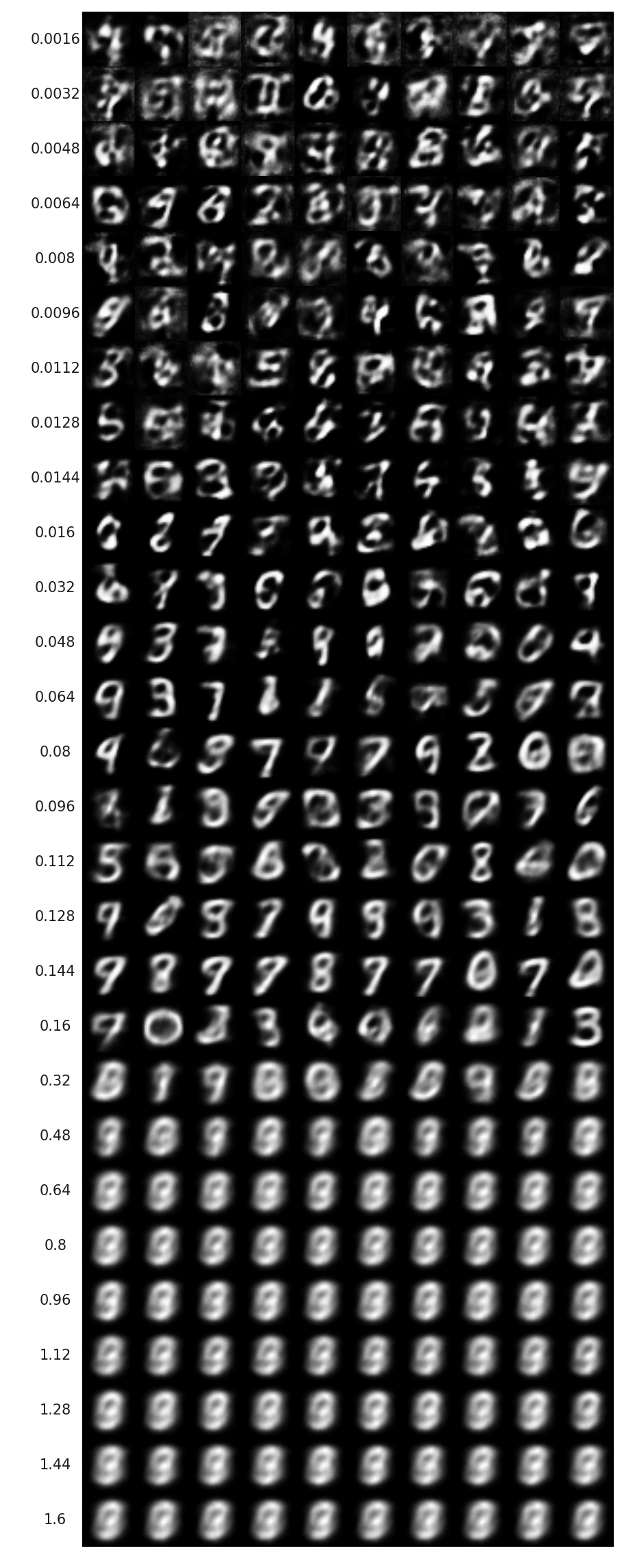}
\end{figure}

\begin{figure}[ht]
\centering
\caption{Mean conditional p-values calculated over 10000 samples generated by the baseline fully-connected models vs. the augmented training fully-connected models (\texttt{gen\_start\_step} = 2400, $p_{\mathrm{sampled}} = 0.125$).}\label{fig:mean_p_values}
\includegraphics[width=0.5\textwidth]{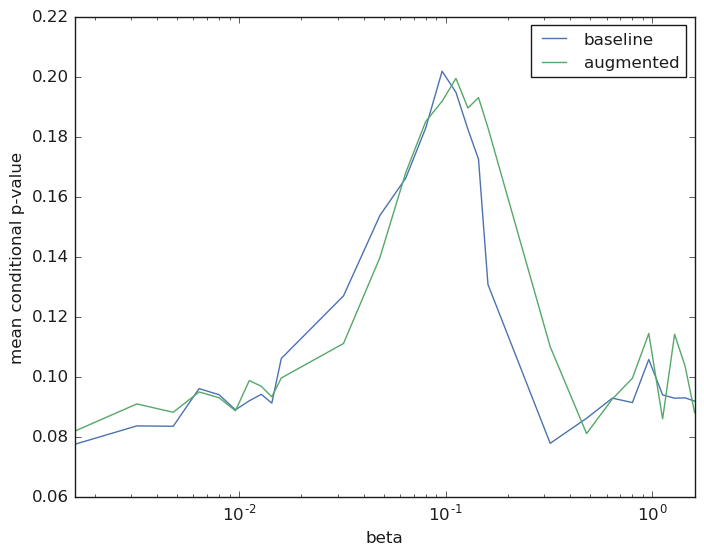}
\end{figure}

\begin{figure}[ht]
\centering
\caption{Median conditional p-values calculated over 10000 samples generated by the baseline fully-connected models vs. the augmented training fully-connected models (\texttt{gen\_start\_step} = 2400, $p_{\mathrm{sampled}} = 0.125$).}\label{fig:median_p_values}
\includegraphics[width=0.5\textwidth]{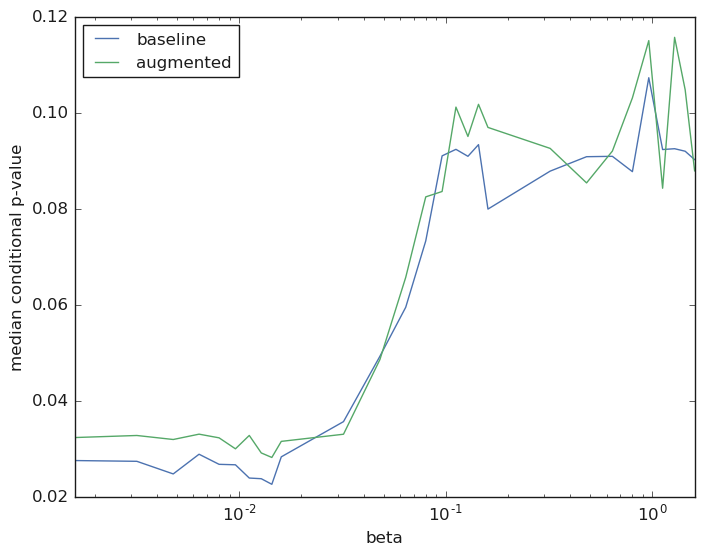}
\end{figure}

We can see that generated loss decreases as the value of $\beta$ decreases over a much narrower range for the fully-connected VAE before it rises again as the model approaches a plain autoencoder in comparison to the deeper convolutional equivalent (Fig~\ref{fig:beta_loss_convol}). Subjectively, however, model with $\beta = 0.016$ still does not seem to be the best generative model (Fig~\ref{fig:mnist_grid}). The fully-connected MNIST VAE overall is a worse generative model so it is not as obvious subjectively, but the observation that repeated encoding and decoding lead to more typical examples of the class still holds (Fig~\ref{fig:0048_repeated_autoencoding},~\ref{fig:0048_train_repeated_autoencoding}, and~\ref{fig:0048_generated_train_repeated_conditional_p_values}). The results of applying augmented training to this fully-connected MNIST VAE are again negative (Fig~\ref{fig:beta_loss_augmented_late_8},~\ref{fig:mnist_grid_augmented_late_8},~\ref{fig:mean_p_values}, and~\ref{fig:median_p_values}) in terms of quantitative measures.

\end{document}